\documentclass[a4paper, 10pt, conference]{ieeeconf}      
\IEEEoverridecommandlockouts                              
\usepackage[utf8]{inputenc} 
\usepackage[T1]{fontenc}    
\usepackage{hyperref}
\hypersetup{colorlinks,allcolors=black}
\usepackage{times}

\usepackage{multicol}

\usepackage{times}
\usepackage{graphicx} 
\usepackage{booktabs}
\usepackage{pifont}
\usepackage{algorithm}
\usepackage[noend]{algpseudocode}

\usepackage{subcaption} 
\usepackage{color} 
\usepackage{amsmath} 
\usepackage{amssymb}  
\usepackage{tikz-cd}
\usepackage{multirow}
\usepackage{booktabs,etoolbox} 
\usepackage{stfloats} 

\usepackage[keeplastbox]{flushend}

\usepackage[style=ieee,natbib=true,backend=bibtex]{biblatex}
\newcommand{\ra}[1]{\renewcommand{\arraystretch}{#1}}

\graphicspath{{figs/}} 

\title{\LARGE \bf
Directional Primitives for Uncertainty-Aware \\Motion Estimation in Urban Environments
}

\author{Ransalu Senanayake, Maneekwan Toyungyernsub, Mingyu Wang, Mykel J. Kochenderfer, and Mac Schwager
\thanks{R. Senanayake, M. Toyungyernsub, and M.J. Kochenderfer are with the Stanford Intelligent Systems Laboratory (SISL); and M. Wang and M. Schwager are with the Stanford Multi-Robot Systems Laboratory (MSL) in the Department of Aeronautics and Astronautics at Stanford University, Stanford, CA 94305, USA (email: \{ransalu, maneekwt, mingyuw, mykel,schwager\}@stanford.edu\}).}
}

\begin{document}

\maketitle
\thispagestyle{empty}
\pagestyle{empty}

\maketitle

\begin{abstract}
We can use driving data collected over a long period of time to extract rich information about how vehicles behave in different areas of the roads. In this paper, we introduce the concept of \emph{directional primitives}, which is a representation of prior information of road networks. Specifically, we represent the uncertainty of directions using a mixture of von Mises distributions and associated speeds using gamma distributions. These location-dependent primitives can be combined with motion information of surrounding vehicles to predict their future behavior in the form of probability distributions. Experiments conducted on highways, intersections, and roundabouts in the Carla simulator, as well as real-world urban driving datasets, indicate that primitives lead to better uncertainty-aware motion estimation. 
\end{abstract}

\section{Introduction}
\label{sec:intro}

Autonomous vehicles will not only have to interact with themselves but also will have to coexist with other vehicles operated by humans. Hence, it is important for autonomous cars to learn the behavior of both human and robotic agents to safely maneuver in busy urban environments. Similar to a human driver, we expect autonomous cars to learn from experience. For instance, when we, as human drivers, approach an intersection, it is possible for us to anticipate how other vehicles in the surrounding would plausibly act. This prediction depends not only on the current behavior but also heavily on our previous experience of observing vehicle interactions in intersections.

\begin{figure}[b!]
    \centering
    \begin{subfigure}[b]{.75\linewidth}
         \centering
         \includegraphics[width=0.9\linewidth]{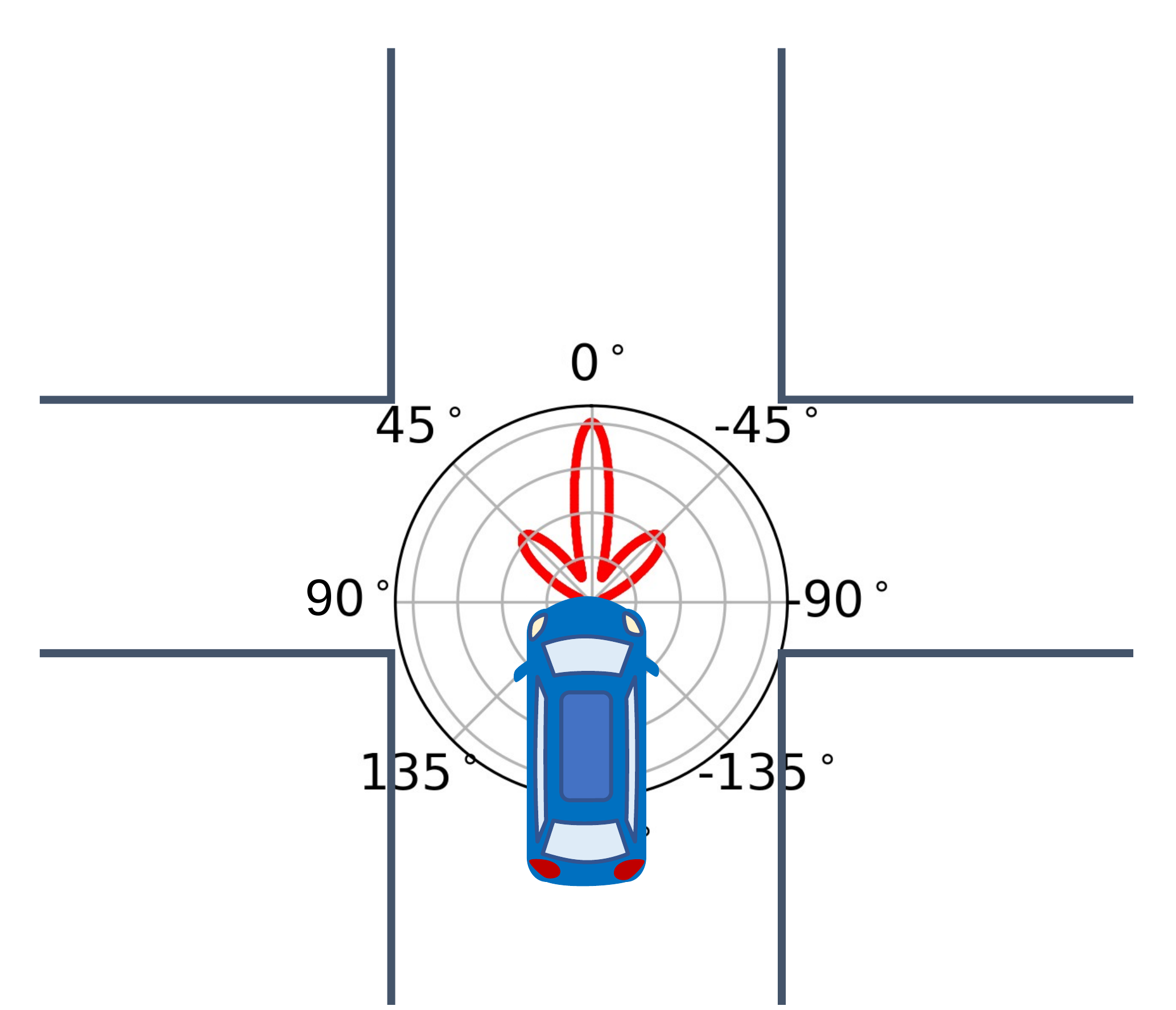}
         \caption{A tri-modal directional distribution obtained from prior vehicle trajectories.}
         \label{fig:motivation_a}
     \end{subfigure}
     \hfill
    \begin{subfigure}[b]{.75\linewidth}
         \centering
         \includegraphics[width=0.7\linewidth]{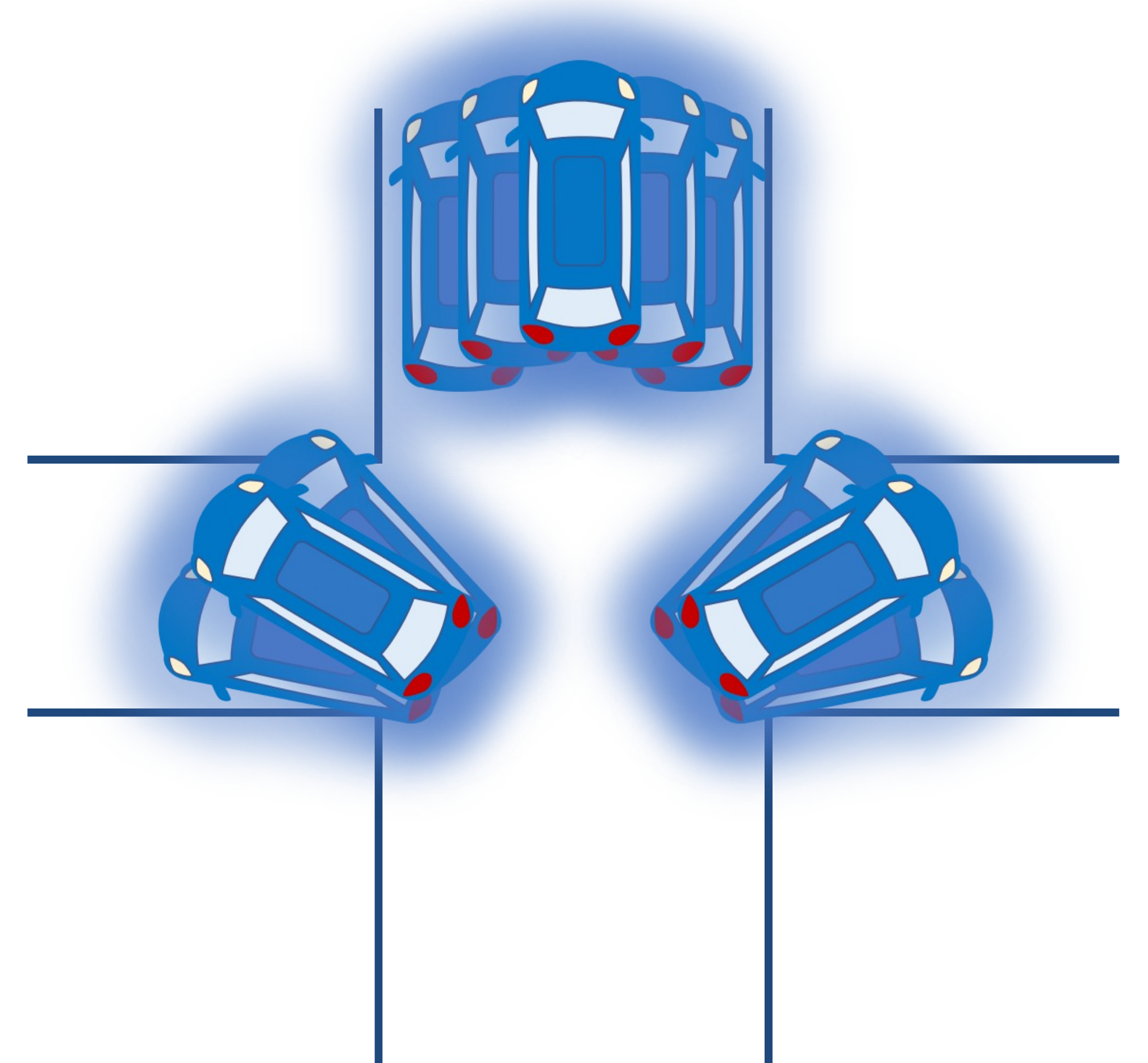}
         \caption{Hallucinated vehicle positions in the next time step.}
         \label{fig:motivation_b}
     \end{subfigure}
    \caption{A vehicle approaches an intersection. For demonstration purposes, we assume one-way roads. The vertical road is the main road and the horizontal roads are collector lanes. Based on prior knowledge, indicated by the polar plot in (a), the vehicle is more likely to continue straight than turn. Some predicted locations are shown in (b).}
    \label{fig:motivation}
\end{figure}

Modern vehicles can be used to collect large amounts of data to gain insight into how drivers typically behave in various parts of the road networks. Going beyond the capacities of human drivers, driverless cars can build lifelong models and can learn from one another. The advancement of vehicle-to-network (V2N) and vehicle-to-everything (V2X) technologies would support the development of such communication needs. 
The behavior of vehicles, both autonomous and human-driven, is highly variable and can be difficult to predict. Therefore, in order to build robust models about the road networks and driver behavior, it is important to consider multimodal probabilistic formulations. For instance, as shown in~Figure~\ref{fig:motivation}, when a car approaches a four-way intersection, there are three possible directions it could turn. It is important that our models capture the uncertainty \cite{shan2019extended} in the future directions. If an autonomous car observes a vehicle ahead, it should be able to infer which directions the vehicle could turn at each position of its possible future trajectories. Such an understanding about the environment can be used to enhance safe decision-making. 

In this paper, we propose {\em directional primitives}, a representation of prior multimodal directional uncertainties that are local to different segments of the road networks. Unlike previous approaches, our focus is to:
\begin{enumerate}
    \item model the \emph{uncertainty} in speed as well as directions,
    \item associate prior knowledge with current knowledge to make accurate predictions, and
    \item use prior information to generate possible trajectories.
\end{enumerate}  

\section{Background}
\label{sec:back}

\subsection{Primitives in robotics}
\label{sec:prim_rob}

The idea of incorporating prior information into likelihood models is widely discussed in Bayesian inference. They have various applications to robotics including simultaneous localization and mapping, Bayesian filtering \cite{fox2003bayesian}, and Bayesian networks for human-machine interaction \cite{tahboub2006intelligent}. Prior information about motion plays an important role in motor control tasks in animals \cite{polit1979characteristics, flash2005motor}. A set of low-level motor skills known as movement primitives \cite{paraschos2013probabilistic} and motion primitives have also been successfully used in robotics applications such as flight control \cite{perk2006motion} and motion planning \cite{lacaze1998path,pivtoraiko2009differentially}. In an imitation learning setting for human-robot interaction, \citeauthor{campbell2019} \cite{campbell2019} learn interactive motor skills using ``interaction primitives.'' The directional primitives we propose in this paper are a set of prior information about possible directions local to a given geographic area. A set of these low-level directional primitives can be used for high-level motion prediction and trajectory generation. 

\subsection{Motion estimation}
\label{sec:mot_est}

Various methods have been used to model the motion of vehicles and pedestrians \cite{dasgupta_cvpr16, park_2018}, predicting occupancy levels \cite{rans_2016,bhm_2017,auto_2018}, and predicting trajectories \cite{park_2018, wilko_2017}. Much of this prior research framed these problems as a supervised learning problem and typically do not take into account prior knowledge about location-specific agent behavior for estimation. 

In contrast with the objective of motion prediction research, our focus is incorporating high-level information about the behavior of observed vehicles into decision-making. For instance, if a vehicle in front of us is signaling to turn right in the intersection shown in Figure~\ref{fig:motivation}, we can specify this current notion of signaling as a directional probability distribution. This distribution can then be used in conjunction with the prior distributions that specify the probabilities of directions and speeds of vehicles in the past in order to infer the possible future locations of the vehicle. 

Estimating the current direction of a moving object has been studied by \citeauthor{senanayake2018directional} \cite{senanayake2018directional}. This work has been extended to the temporal domain using LSTMs \cite{zhi2019klstm}. While our proposed model can plausibly be extended to the temporal domain \cite{vintr2019time,vintr2019spatio}, this paper focuses only on the spatial domain. None of these previous methods consider the variation of speed, which is also crucial for accurate motion prediction.

\section{Directional primitives}
\label{sec:dir_prim}

\subsection{Location-dependent direction-speed priors}
\label{sec:ds_prior}
We model the uncertainty of directions at every location in a road network. The uncertainty of speeds associated with each mode of direction is also modeled.

\subsubsection{Uncertainty of directions}
\label{sec:d_prior}

\begin{figure}[t]
    \centering
    \begin{subfigure}[b]{.48\linewidth}
         \centering
         \includegraphics[width=1.\linewidth]{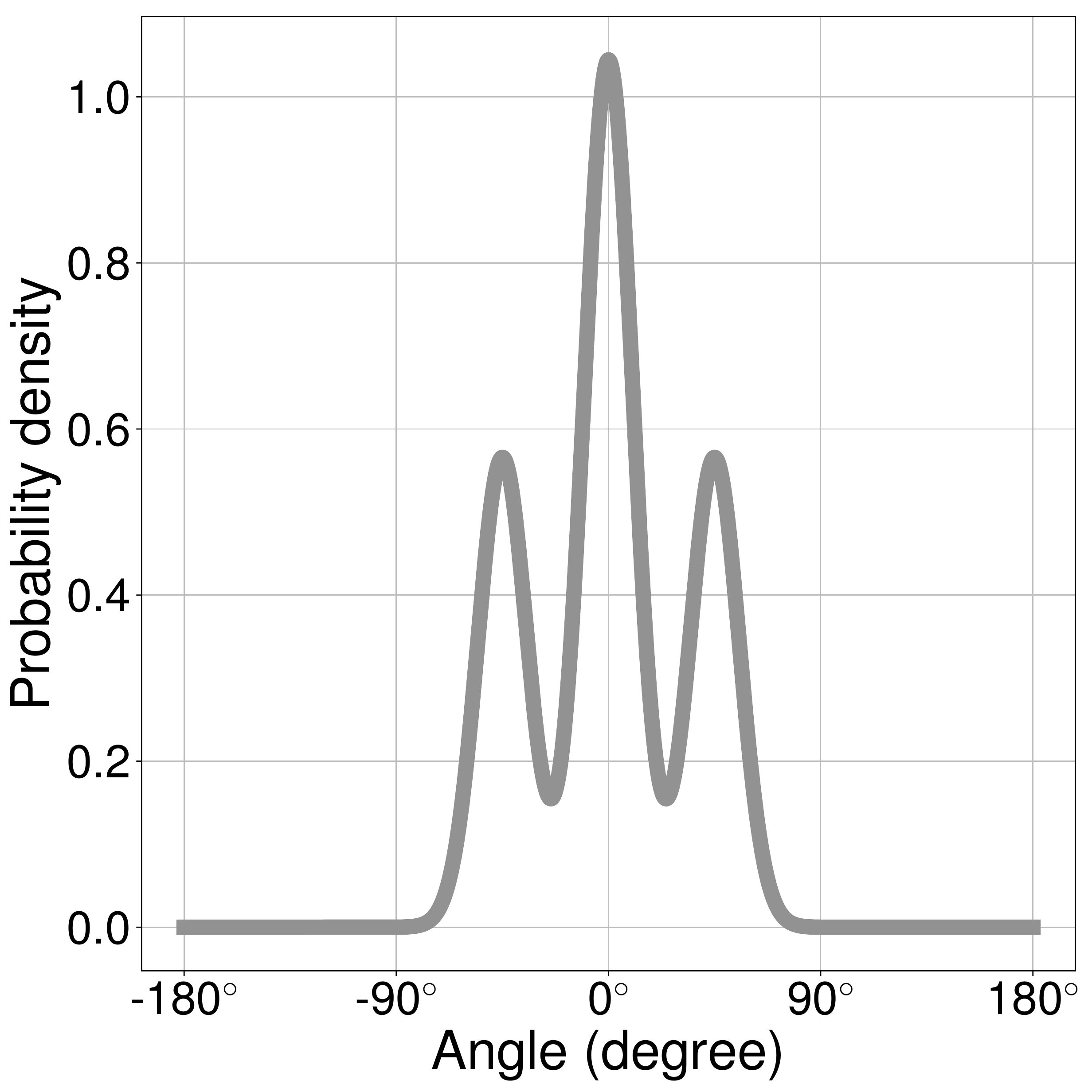}
         \caption{Linear projection}
         \label{fig:linear}
     \end{subfigure}
     \hfill
    \begin{subfigure}[b]{.48\linewidth}
         \includegraphics[width=1.\linewidth]{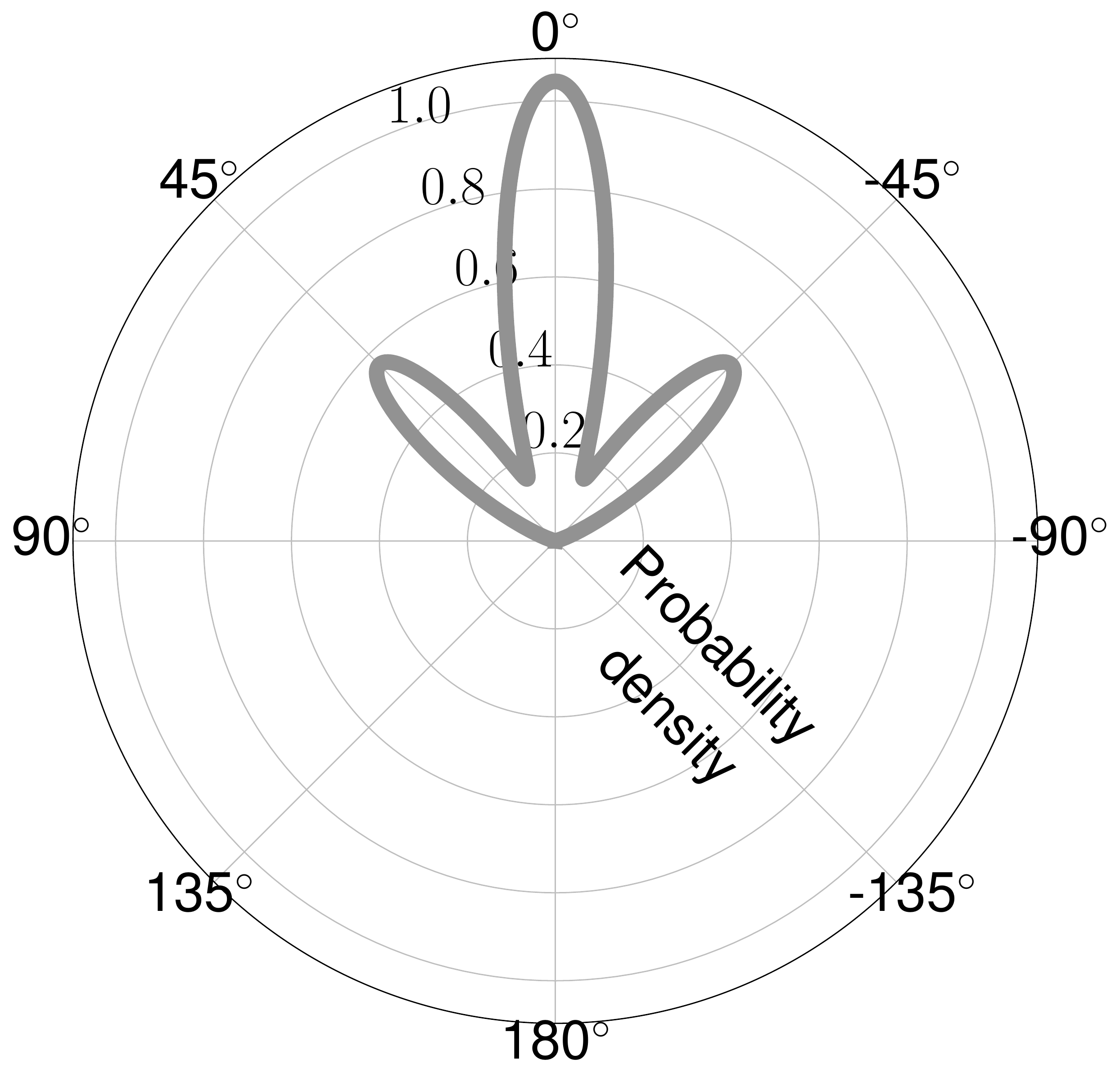}
         \caption{Polar projection}
         \label{fig:polar}
     \end{subfigure}
    \caption{Two different representations (projections) for a mixture of von Mises distributions with parameters $M=3, \boldsymbol\mu=[-45^\circ, 0^\circ, 45^\circ], \boldsymbol\kappa=[20,20,20], \text{ and } \mathbf{w}=[0.25,0.5,0.25]$.}
    \label{fig:vm_intro}
\end{figure}

In order to effectively model uncertainty in real-world robotic applications such as autonomous driving, the stochasticity of directions must be modeled. Unlike many other physical quantities, there are several challenges when modeling directional variability using a probability density function. A density function that represents directions must have a limited support $\theta \in [0,2\pi)$ with the two limits, $0$ and $2\pi$, being the same.

Circular distributions such as Bingham distributions have been used in filtering problems \cite{kurz2014recursive,gilitschenski2015unscented}. Such distributions characterized by rotation matrices are ideal for modeling joint angles in $SE(3)$. In this paper, we consider the von Mises distribution \cite{MardiaEdw1982,DirectStat} used for directional grid maps (DGM) \cite{senanayake2018directional}. Intuitively, a von Mises distribution can be thought of as a Gaussian distribution wrapped around a circle. 

In order to understand in which directions vehicles typically move in different segments of the road network, we begin by discretizing the environment into a grid with $P$ cells. A mixture of von Mises distributions is assigned to each cell. The mixture is to handle the multiple possible directions \cite{IvanovicSchmerlingEtAl2018, senanayake2018directional}. This mixture,  
\begin{equation}
\label{vm_mix}
    p(\theta \mid \mathbf{x}) = \sum_{m=1}^M w_m \mathcal{VM}(\theta;\mu_m, \kappa_m),
\end{equation}
models the probability of possible directions $\theta \in [0,2\pi)$ at anywhere in the longitude-latitude space $\mathbf{x} \in \mathbb{R}^2$. With weights $\sum_{m=1}^M w_m = 1$, the distribution is composed of $M$ von Mises distributions,
\begin{equation}
    \mathcal{VM}(\theta;\mu_m, \kappa_m) = \frac{1}{2\pi J_0(\kappa_m)} \exp\big(\kappa_m \cos(\theta - \mu_m) \big) ,
\end{equation}
where $\mu_m$ is the mean direction and $\kappa_m$ is the concentration.\footnote{Analogous to a Gaussian distribution, these can be thought as the mean and inverse variance parameters.} The higher the concentration parameter, the lower the dispersion of angles is. $J_r(\kappa)$ is the $r$th order and $1$st kind modified Bessel function. The mean and variance of the $m$th von Mises distribution are $\mu_m$ and $\big(1-J_1(\kappa_m)/J_0(\kappa_m) \big)$, respectively. Figure~\ref{fig:vm_intro} shows a mixture of von Mises distributions with three components, also known as modes. 

For the entire road network with $P$ cells and $M_p$ mixture components for the $p$th cell, the set of parameters is $\{\{( \mu_{pm}, \kappa_{pm}, w_{pm}) \}_{m=1}^{M_p} \}_{p=1}^P$. With angles extracted from vehicle trajectory data, these parameters can be learned using $P$ individual expectation-maximization routines \cite{ester1996density,senanayake2018directional}. $M_p$ is determined for each cell using a density-based clustering algorithm \cite{ester1996density}.

\subsubsection{Uncertainty of speeds}
\label{sec:s_prior}

In Section~\ref{sec:d_prior}, we obtained a multimodal distribution of directions. In addition to the directions, we also want to model the distribution of speed for each directional modality. Having such speed priors is important because the speeds are different in various segments of the road network. For instance, the speed of vehicles is relatively low near intersections, crosswalks, and roundabouts compared to a highway. Since speed is strictly a non-negative quantity, it is modeled using a gamma distribution given by,
\begin{equation}
    p(s \mid \mathbf{x}) = \frac{\beta^\alpha}{\Gamma(\alpha)} s^{\alpha-1} \exp{(-\beta s)},
\end{equation}
where $\alpha$ and $\beta$ are the shape and rate parameters, respectively. $\Gamma$ is the gamma function. The mean and variance of a gamma distribution can be computed as $\alpha \beta^{-1}$ and $\alpha \beta^{-2}$, respectively. Gamma distributions are estimated by maximizing the likelihood \cite{anderson1975improved} with data that lie within twice the standard deviation of each von Mises component. 

\subsection{Hallucinating future locations}
\label{sec:hall}

\begin{figure*}[t]
    \centering
    \begin{subfigure}[b]{.3\linewidth}
         \centering
         \includegraphics[width=\textwidth]{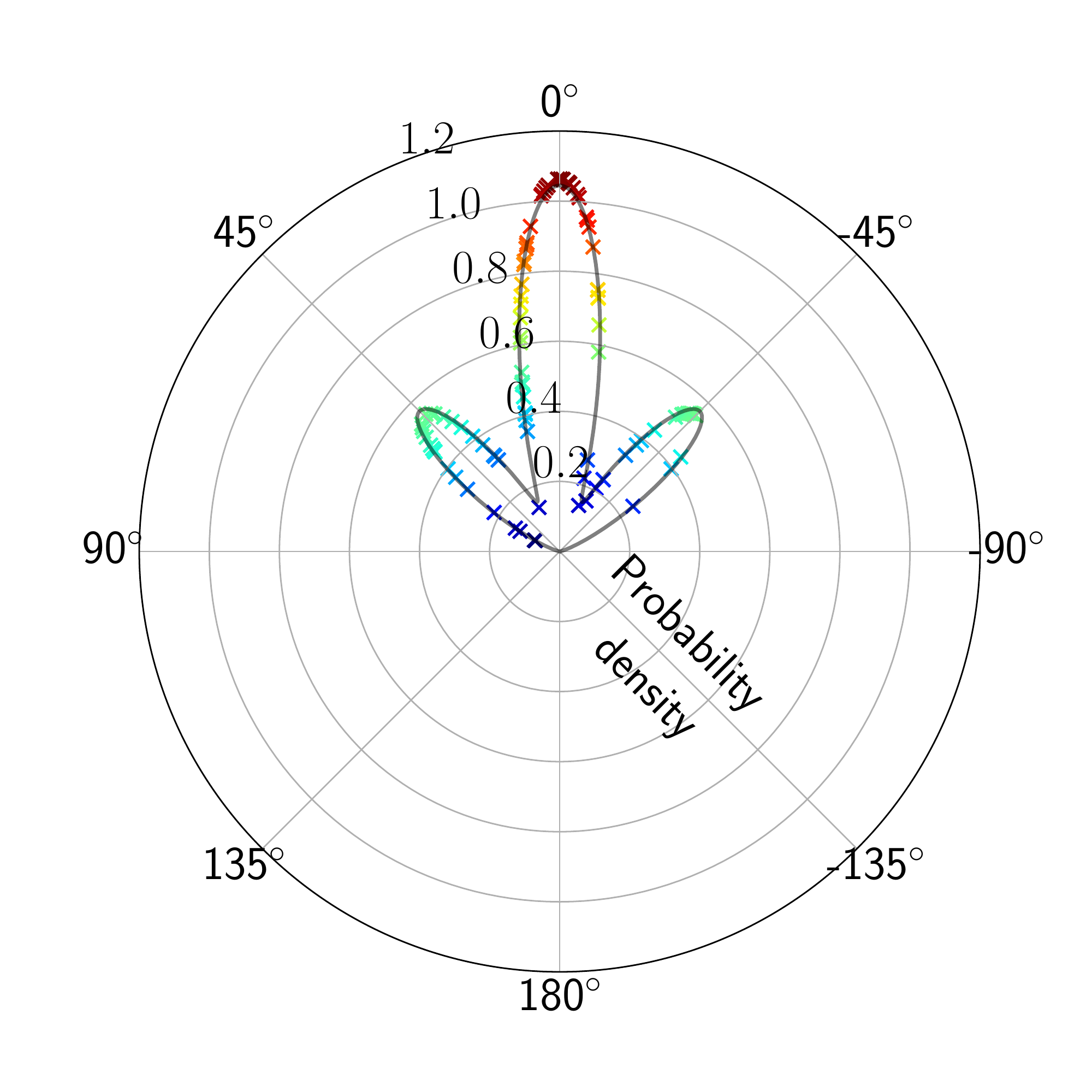}
         \caption{Sampling the directional distribution}
         \label{fig:sam}
     \end{subfigure}
     \hfill
    \begin{subfigure}[b]{.3\linewidth}
        \includegraphics[width=\textwidth]{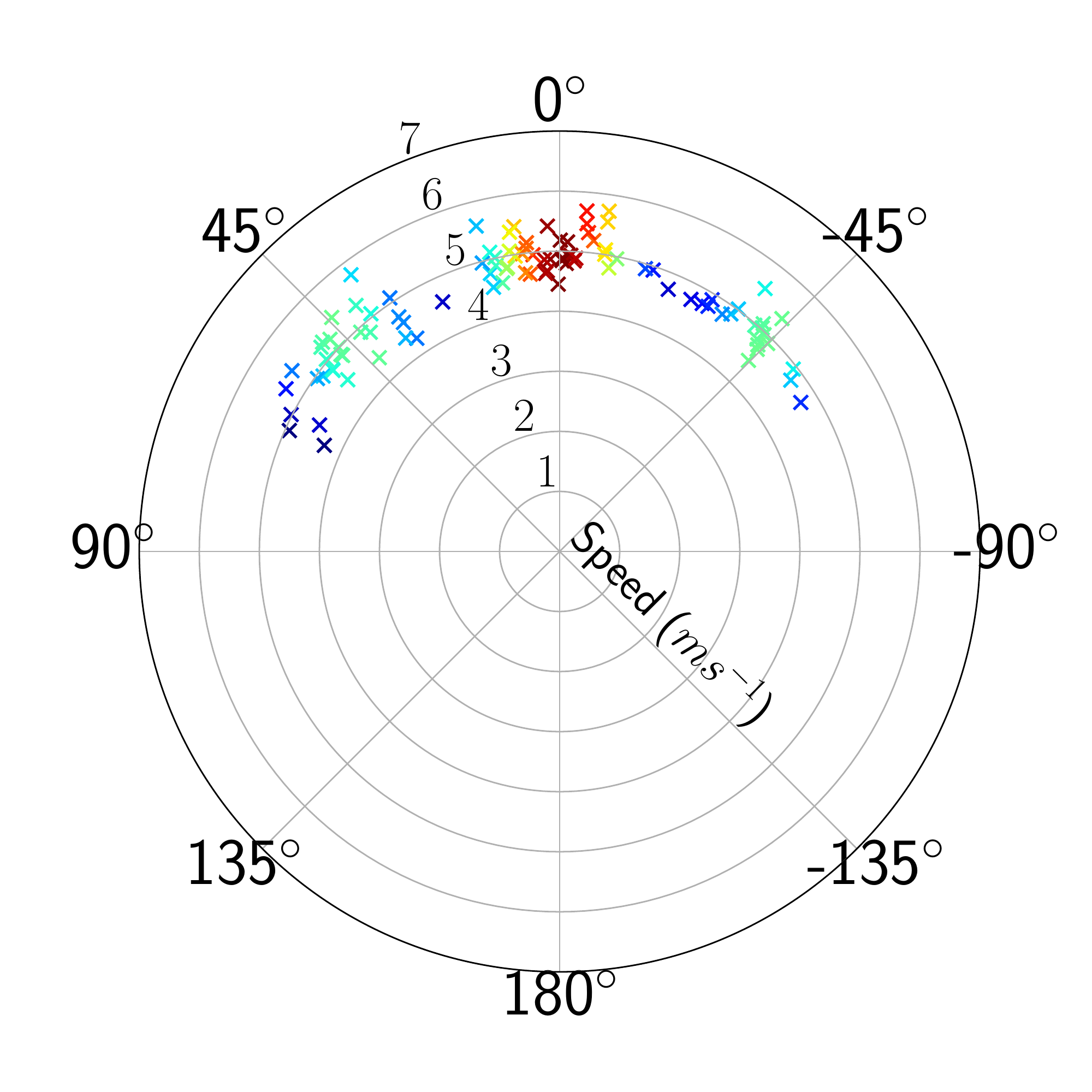}
         \caption{Combining the samples with speeds}
         \label{fig:polar2}
     \end{subfigure}
     \hfill
    \begin{subfigure}[b]{.33\linewidth}
        \includegraphics[width=\textwidth]{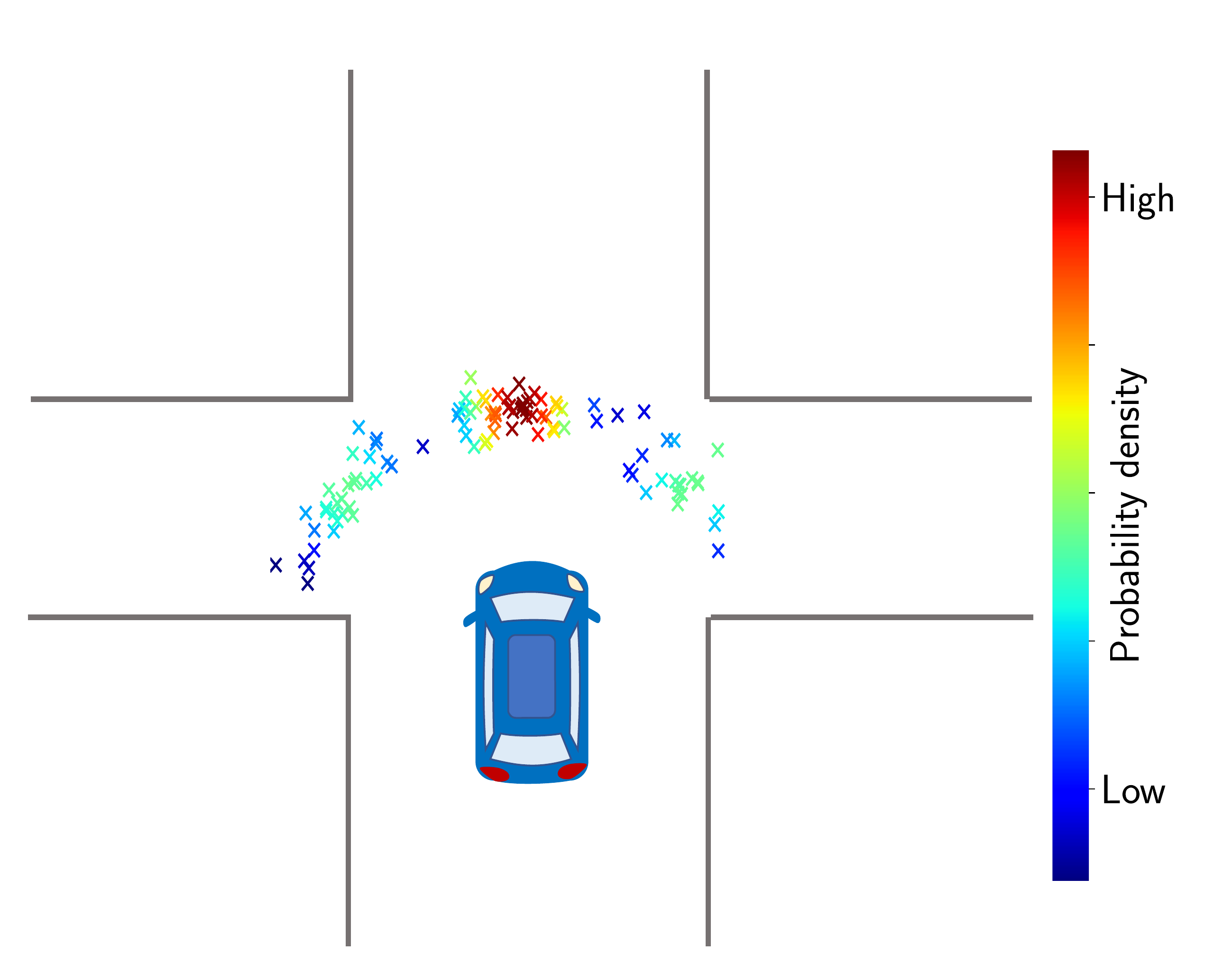}
         \caption{Projected positions}
         \label{fig:polar3}
    \end{subfigure}
    \caption{(a) A mixture of von Mises distribution and 100 samples drawn from it. (b) With different speeds in different directions, samples are projected into the future. (c) Projected locations of the vehicle.}
    \label{fig:angle_pos}
\end{figure*}

If an autonomous car observes another vehicle at a particular location, then the corresponding cell the vehicle belongs to can be used to predict where the observed vehicle could move. As depicted in Figure~\ref{fig:motivation}, finitely many such hallucinations can be obtained by sampling the possible directions from the mixture of von Mises distribution and projecting the vehicle towards the sampled directions of motion.

To sample from the mixture, we first sample $\tilde{m}_p \in \{1, 2, \dots, M_p \}$ from a categorical distribution $\mathcal{CAT}([w_{p1}, w_{p2}, \dots, w_{pM_p}])$. Then, we sample an angle $\tilde{\theta}_p$ from the $\tilde{m}_p$th von Mises mixture component \cite{ulrich1984computer,s-vae18}. This process can be repeated to obtain $N$ samples. If a sample has a speed $s$, in a unit time, the vehicle will be projected into the positions $\{ [s\cos(\tilde{\theta}_p^{(n)}), s\sin(\tilde{\theta}_p^{(n)})] \}_{n=1}^N$ in the vehicle's coordinate system. If the distribution of speeds corresponding to the particular location is also known, it can also be incorporated into computations by replacing $s$ with $s(\tilde{\theta}_p^{(n)})$ and sampling from the gamma distribution introduced in Section~\ref{sec:s_prior}. As illustrated in Figure~\ref{fig:angle_pos}, uncertainty in directions and speeds would project the vehicle not only into various angles, but also into different positions.

\subsection{Incorporating current knowledge}
\label{sec:curr_know}

\begin{figure}[]
    \centering
    \includegraphics[width=0.65\linewidth]{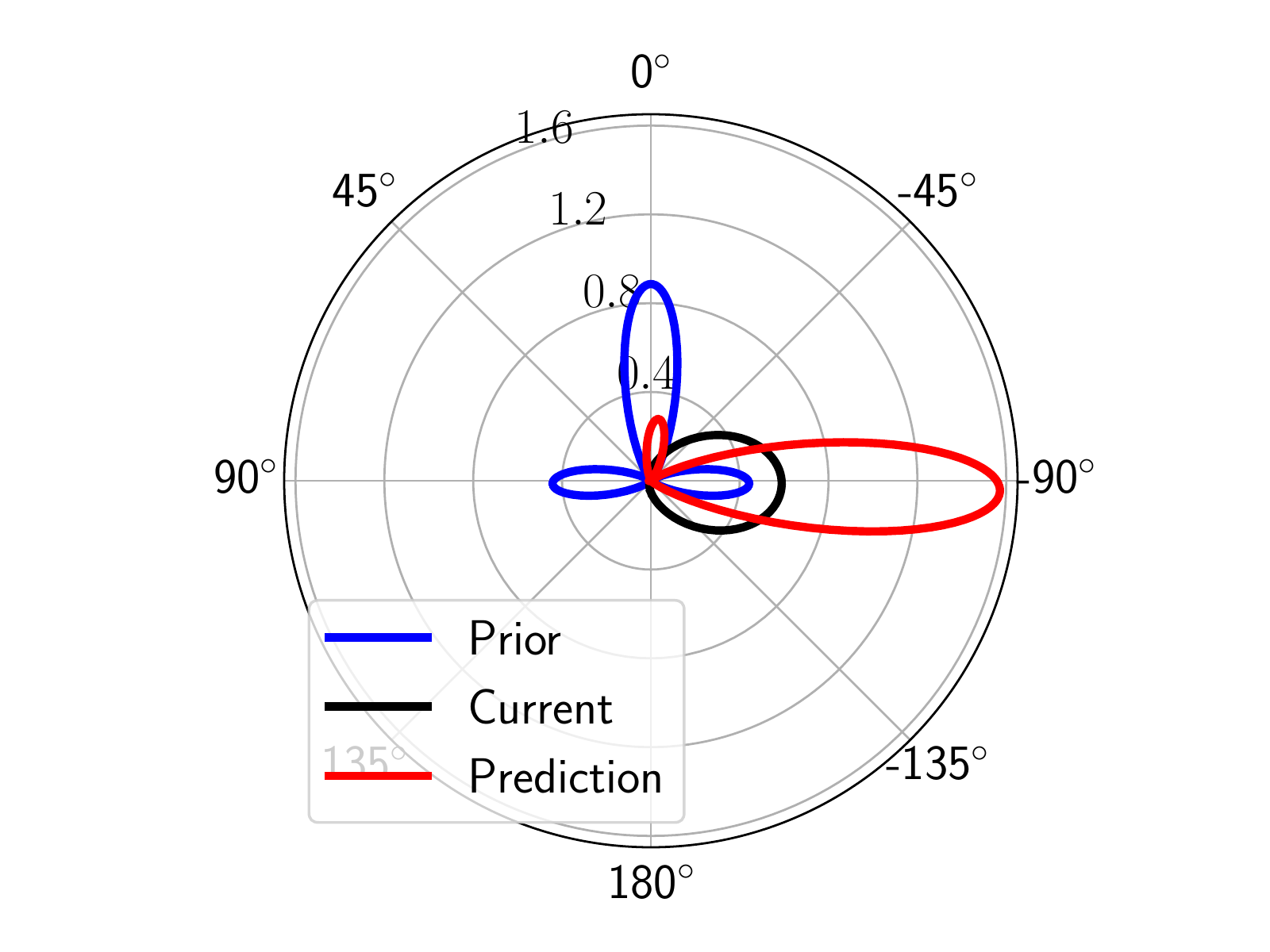}
    \caption{The prior distribution $p_0(\theta)$ obtained from past vehicle trajectories is shown in blue. The distribution in black $p_t(\theta)=\mathcal{VM}(-90^\circ,2.5)$ indicates the belief about a vehicle. For instance, if the vehicle signals to turn right, we can set the mean parameter to $-90^\circ$ and concentration to a very low value (i.e. high uncertainty). Combining the prior and current information, we can obtain a new predictive distribution shown in red. }
    \label{fig:prod}
\end{figure}

Note that the estimation in Section~\ref{sec:hall} is purely based on previous experiences of observing the behavior of thousands of vehicles in the past. Nonetheless, past information alone is not sufficient for accurate decision-making. Within our framework, we can also effectively use high-level information about the environment to bias the predictions so as to make informed decisions.

If we have some indications of how an observed car would behave, then we incorporate this current high-level information with the prior information to make better predictions. For instance, consider the distributions in Figure~\ref{fig:prod}. According to the prior information $p_0(\theta)$, shown in blue, there is a very high chance that the vehicle would continue straight, and there is less chance that it would turn either left or right. However, if we observe the vehicle leans towards the right lane or it signals the right blinkers, we have a current probability distribution $p_t(\theta)$ with the mean direction towards the right and very low concentration indicating high uncertainty.

\begin{figure}[b]
    \centering
     \includegraphics[width=\linewidth]{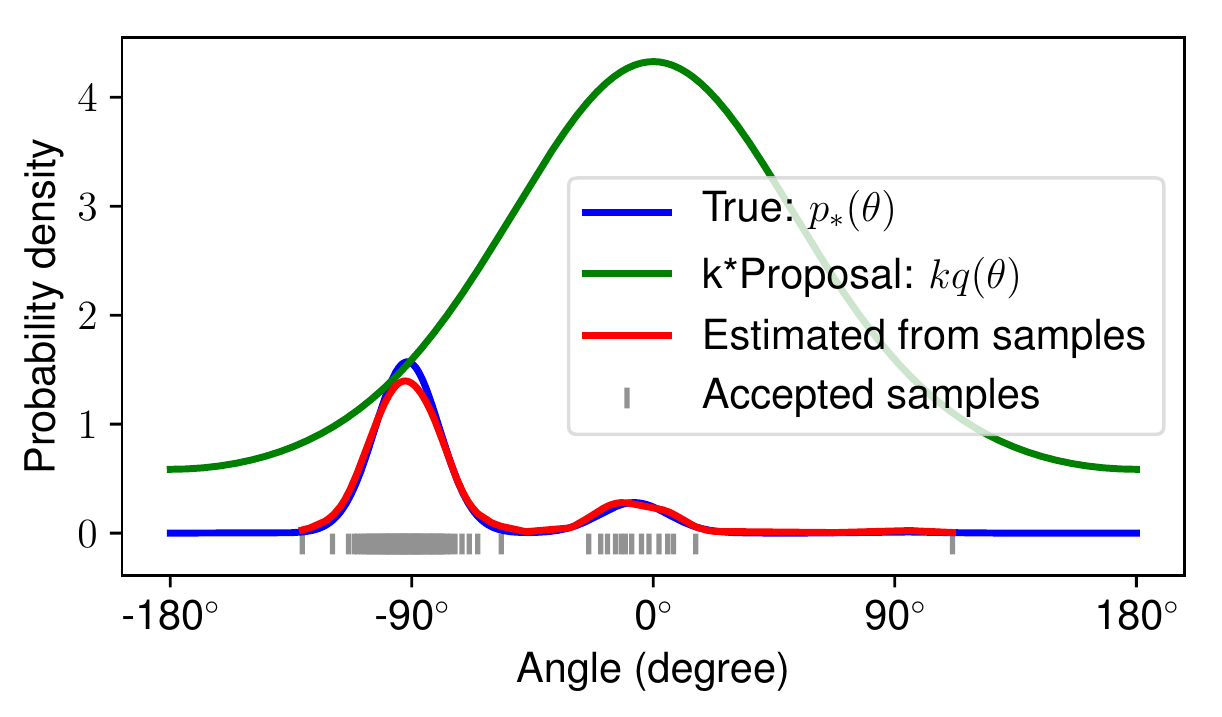}
    \caption{Rejection sampling for a mixture of von Mises distribution. Kernel density estimate \cite{davis2011remarks} on the accepted samples was used to plot the red curve.}
    \label{fig:rej}
\end{figure}

Based on the prior and current information, we can compute the new directional distribution as,
\begin{equation}
    \label{eq:post}
    p_*(\theta) \propto p_t(\theta) p_0(\theta).
\end{equation}

Because of current information, the new distribution $p_*(\theta)$, shown in red in Figure~\ref{fig:prod}, has a higher probability towards right and negligible probability towards straight and almost zero probability towards left.  

In order to obtain samples from the ``product of mixtures'' $p_*(\theta)$, a Monte Carlo technique \cite{glynn1989importance} can be used. In this paper, as illustrated in Figure~\ref{fig:rej}, we propose a \emph{rejection sampling} scheme \cite{casella2004generalized}. Firstly, we select an arbitrarily large constant $k$ and a unimodal von Mises distribution with a small concentration as the \emph{proposal distribution} $q(\theta)$. This distribution should be broad enough to enclose the underlying true distribution. We then draw a sample $\tilde{\theta}$ from $q(\theta)$ and evaluate $k \times q(\tilde{\theta})$ and $p_*(\tilde{\theta})$. Independent to this process, another sample $\tilde{u}$ is drawn from the uniform distribution $\mathcal{U}[0,k*q(\tilde{\theta})]$. Samples $\tilde{\theta}$ are accepted as samples from $p_*(\theta)$, if $\tilde{u} \leq p_*(\tilde{\theta})$. The entire process is repeated until it satisfactorily converges \cite{casella2002statistical}. All accepted samples are kept as an approximation to $p_*(\theta)$. In order to further improve the sampling efficiency, similar to sampling from a product of mixture of Gaussians, Gibbs sampling can be performed using a KD-tree \cite{ihler2004efficient}.

\subsection{Generating multi-modal trajectories}
If a vehicle is observed in cell $\mathcal{C}_p$ at time $t$, it is possible to sample an angle $\tilde{\theta}_p^{[t]}$ and move in that direction for one time step. Then we can sample from the directional (and speed) distribution in the new cell. We can repeat this for $T$ time steps. This process is summarized in Algorithm~\ref{algo:traj}.

\begin{algorithm}[]
\caption{Trajectory generation \label{algo:traj}}
\begin{algorithmic}[1]
\State \textbf{input} Initial vehicle position $\mathbf{x}_0$
\For{$k \in \{0,1,\ldots,K\}$ trajectories}
    \For{$t \in \{0,1,\ldots,T_k\}$ time steps}
        \State $\mathcal{C}_p$ $\leftarrow$ Determine the cell corresponding to $\mathbf{x}_t^{[k]}$ 
        \State $\tilde{\theta}_p^{[t,k]}$ $\leftarrow$ Sample from the distribution over $\mathcal{C}_p$  
        \State $\mathbf{x}_{t+1}^{[k]}$ $\leftarrow$ Move in the direction $\tilde{\theta}_p^{[t,k]}$ 
    \EndFor
\EndFor
\State \Return{Trajectories $\{[\mathbf{x}_0^{[k]}, \mathbf{x}_{1}^{[k]}, \ldots, \mathbf{x}_{T_k}^{[k]}] \}_{k=1}^K$ }
\end{algorithmic}
\end{algorithm}

\section{Experiments}

\begin{figure}[b]
    \centering
    \includegraphics[width=0.43\linewidth, height=0.42\linewidth]{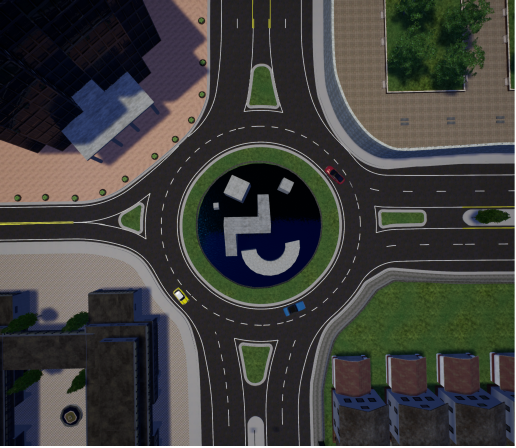}
    \includegraphics[width=0.45\linewidth]{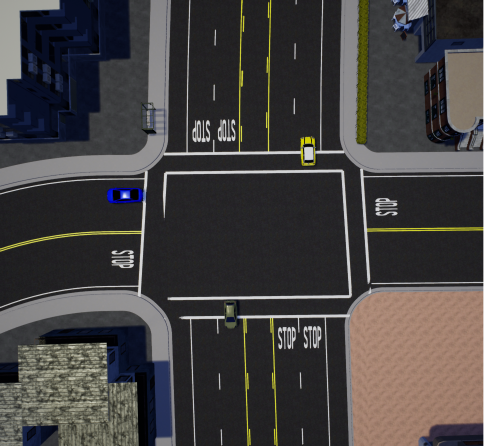}
    \caption{Carla roundabout and intersection.}
    \label{fig:carla1}
\end{figure}

The Carla simulator \cite{dosovitskiy2017carla} was used to collect simulation data. Town 3 of the simulator was used with 20 cars running in the autopilot mode for around 2 hours. To increase diversity, the system was restarted by spawning new cars every 10000 timesteps. Spurious trajectories such as rare collisions that are generated due to the limitations of the driver model of the simulator were removed based on visual inspection. To demonstrate the idea of directional primitives, we considered two important areas of the environment: a roundabout and an intersection (Figure~\ref{fig:carla1}). Stanford Drone dataset \cite{robicquet2016learning} and the Lankershim segment of the NGSIM dataset \cite{alexiadis2004next} were used as real-world datasets. The former is an aerial dataset that contains trajectories of pedestrians, cyclists, etc. whereas the latter contains trajectories of highways in the US.

\begin{figure*}[]
    \centering
    \includegraphics[width=\linewidth]{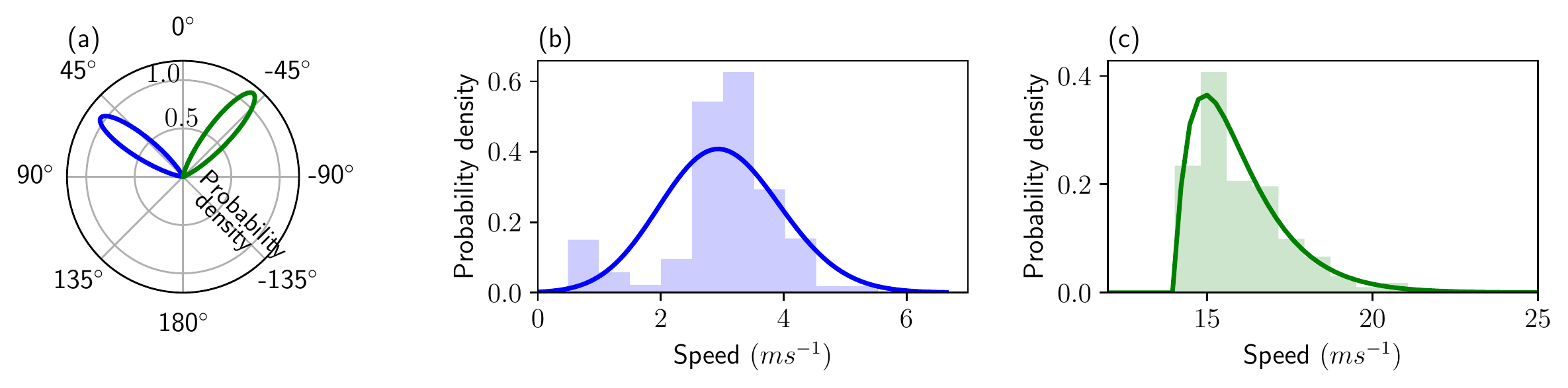}
    \caption{(a) A bimodal directional distribution. Blue and green indicates the two modes. (b) and (c) show the speed distributions, gamma,  corresponding to the blue and red lobes of the directional distribution.}
    \label{fig:gamma_dir}
\end{figure*}

\begin{figure*}[]
    \centering
    \begin{subfigure}[b]{1.\linewidth}
         \centering
         \includegraphics[width=0.36\linewidth]{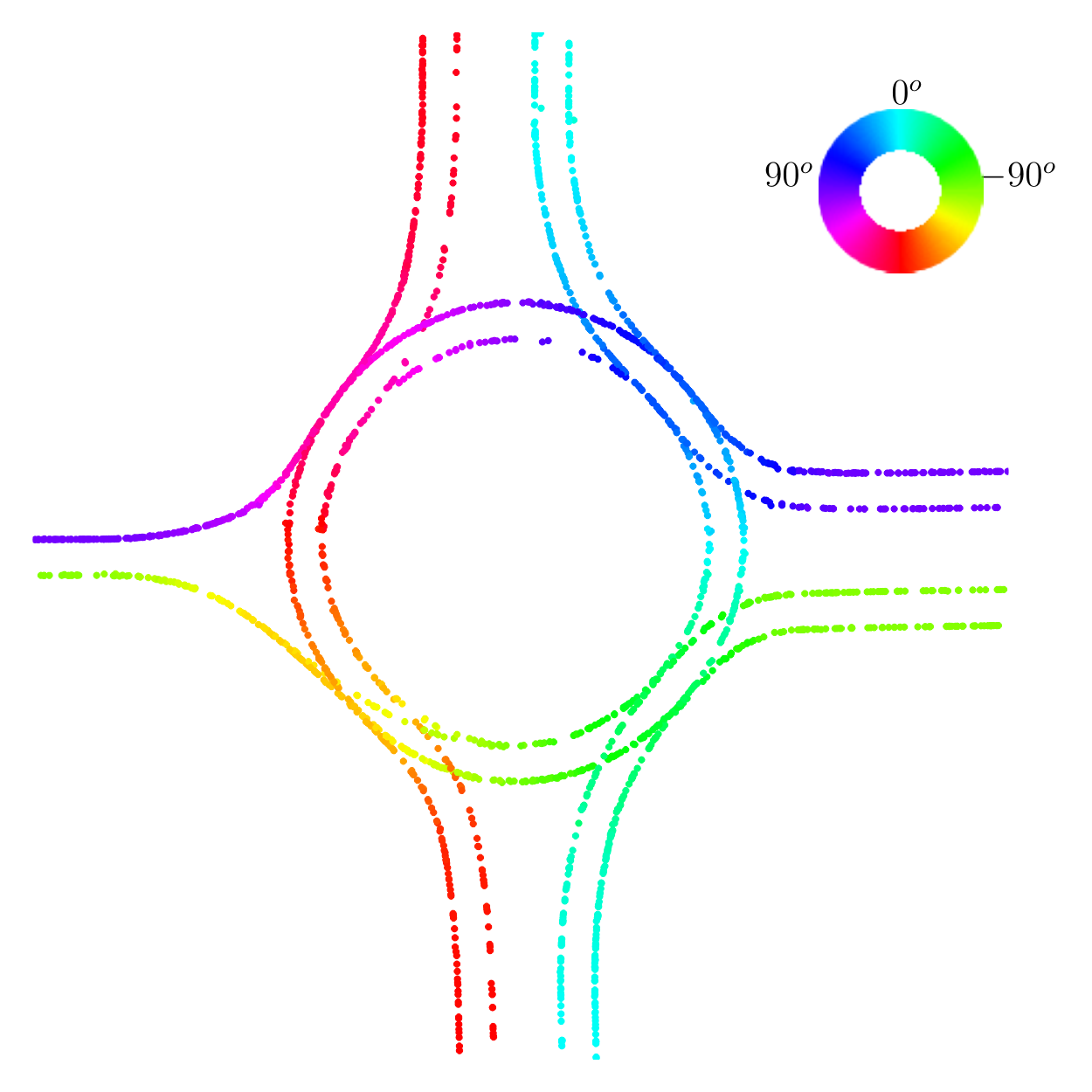}
         \includegraphics[width=0.36\linewidth]{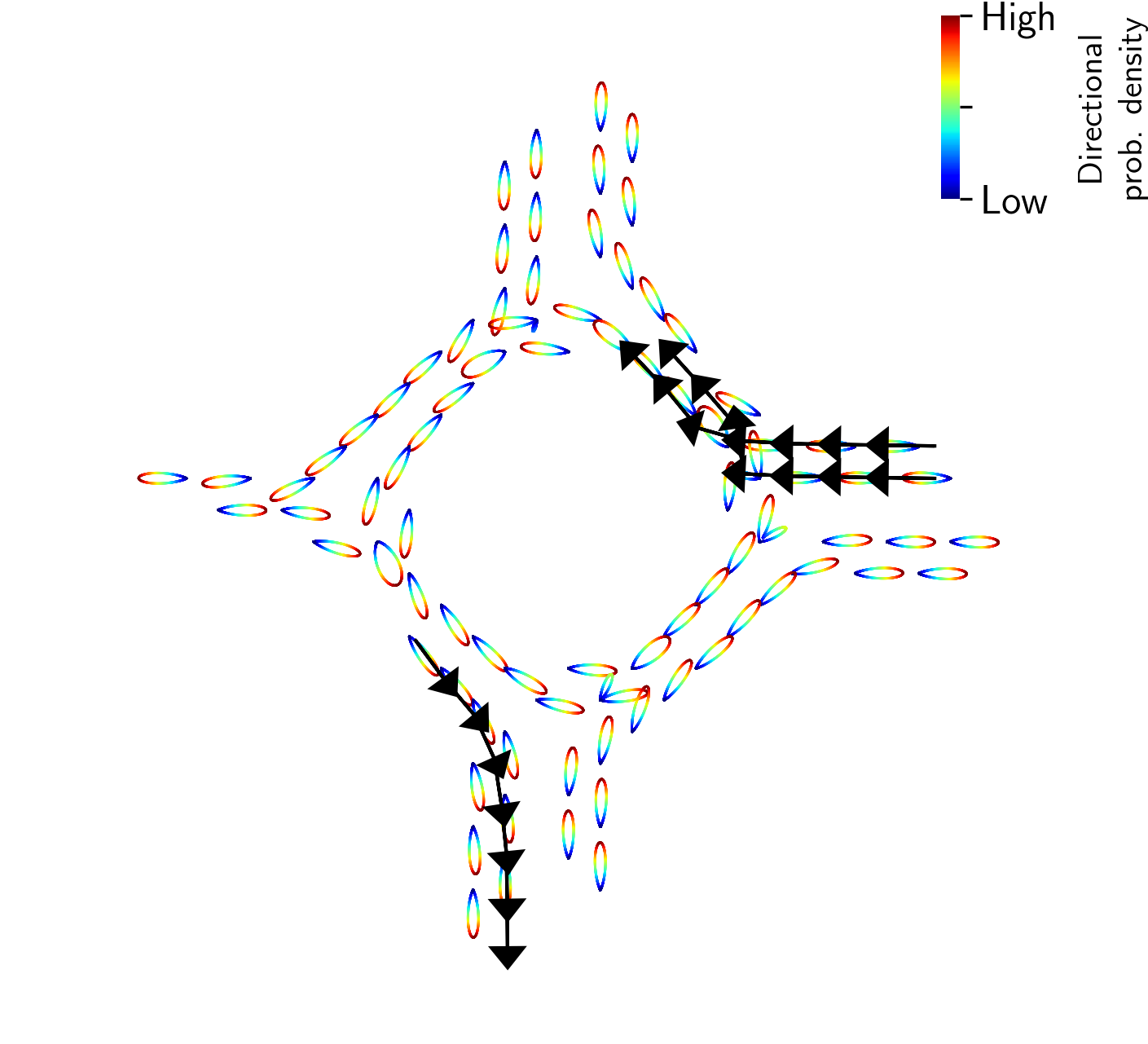}
         \caption{Carla Roundabout}
         \label{fig:carla2_a}
     \end{subfigure}
     \hfill
    \begin{subfigure}[b]{1.\linewidth}
         \centering
        \includegraphics[width=0.36\linewidth]{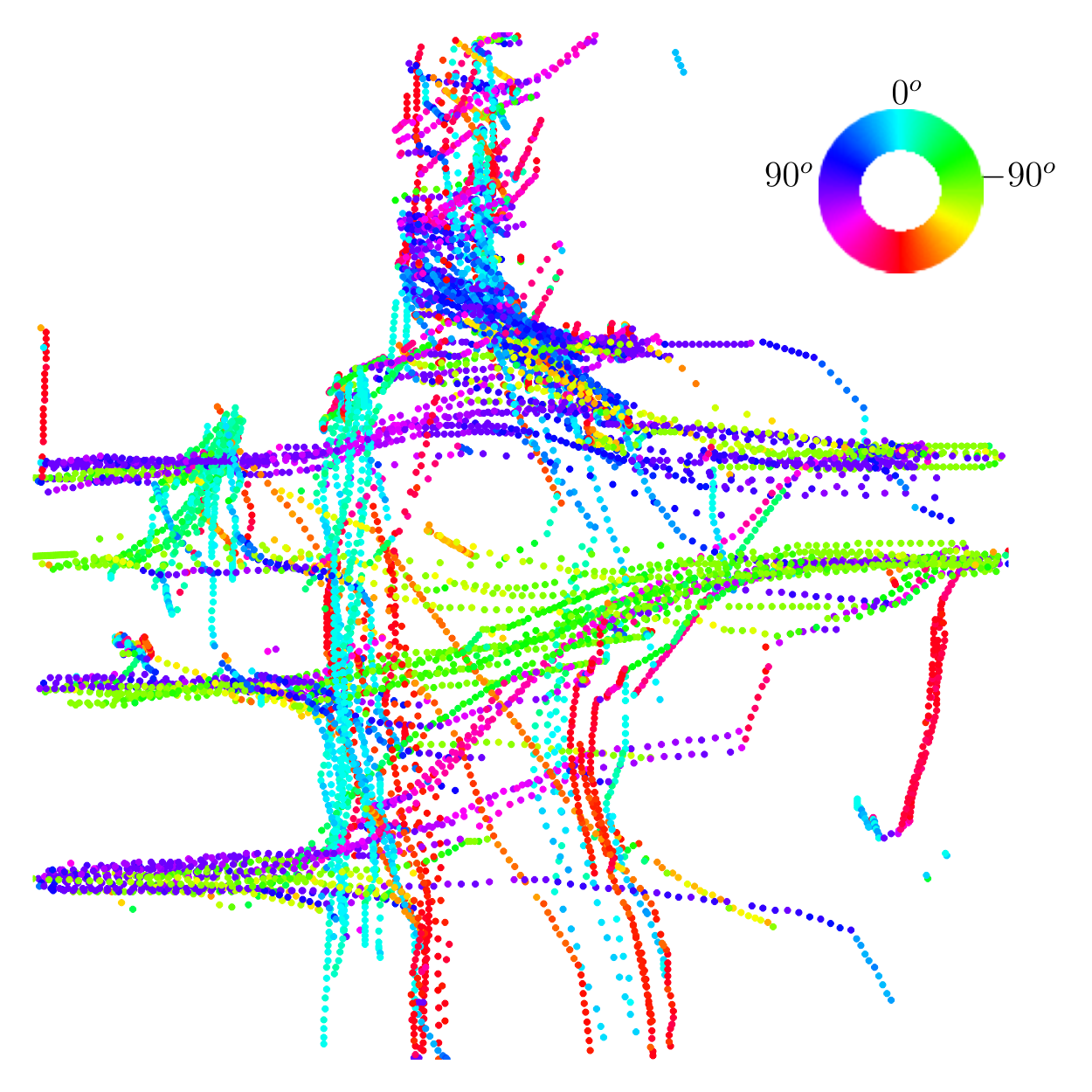}
         \includegraphics[width=0.36\linewidth]{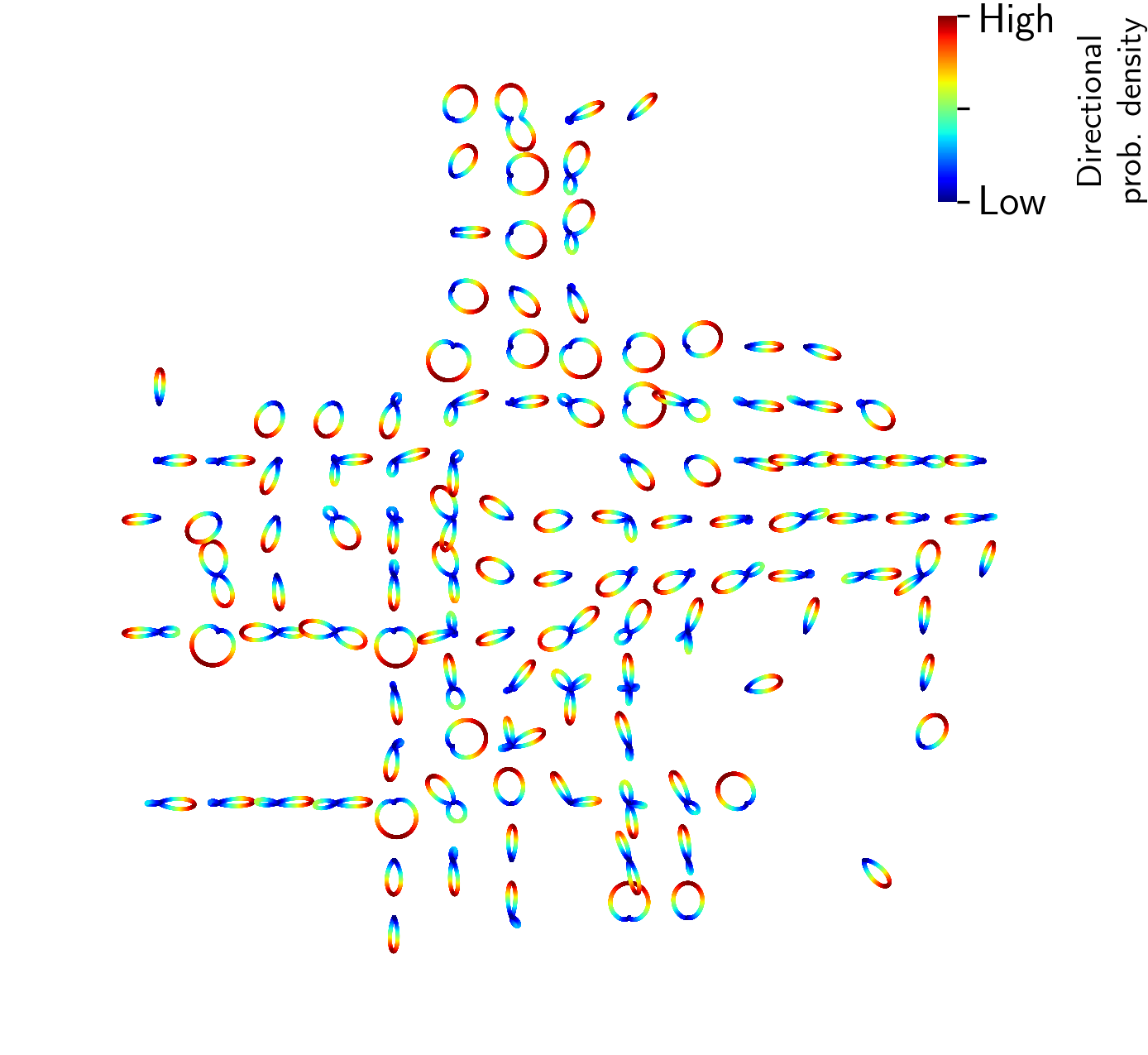}
         \caption{Stanford Death Circle}
         \label{fig:carla2_b}
     \end{subfigure}
    \caption{Directional polar plots for various locations in the (a) Carla roundabout and (b) Stanford Death Circle. The dataset in (a) only contains trajectories of cars while dataset (b) contains trajectories of pedestrians, bicyclists, skateboarders, buses, etc. This plot is obtained by overlaying many individual multimodal directional distributions. Each distribution takes the form of Figure~\ref{fig:motivation_a}. Blue to red indicates lower to higher directional density. Therefore, directions with red are the most probable directions of motion. Black arrows indicate some generated paths starting at three different locations. It is possible to generate finitely many trajectories that originate from anywhere in the environment.}
    \label{fig:carla2}
\end{figure*}

Roads in each environment were divided into cells depending on the road geometry. Von Mises and gamma distributions were then fitted to each cell as described in Sections \ref{sec:d_prior} and \ref{sec:s_prior}, respectively. As shown in Figure~\ref{fig:carla2}, this results in a collection of von Misses distributions spread throughout the environment. For each directional mode in each cell, we also have an associated speed distribution. An example of a pair of speed distributions corresponding to a given cell with a bimodal von Mises directional distribution is shown in Figure~\ref{fig:gamma_dir}.

\begin{table*}[t]
  \centering
  \ra{1}
  \caption{Average probability density for different models. The higher, the better.}
    \begin{tabular}{@{}lccccccc@{}}
    \toprule
    \multicolumn{1}{c}{\bf Dataset} & \phantom{abc} & \multicolumn{4}{c}{\bf Direction} & \phantom{abc} & \multicolumn{1}{c}{\bf Speed} \\
    \cmidrule{3-6} \cmidrule{8-8}
    & & Primitive & DGM & GP & Uninformative & & Primitive \\
    \midrule
    Carla roundabout & &  1.891 $\pm$ 0.471  &  1.480 $\pm$ 0.727 &  1.484 $\pm$ 1.029 &  0.159 $\pm$ 0.000  & & 0.304 $\pm$ 0.126 \\
    Carla intersection &&  1.893 $\pm$ 0.468  &  1.588 $\pm$ 0.758  &  0.657 $\pm$ 0.505 &  0.159 $\pm$ 0.000  & &  0.269 $\pm$ 0.110 \\
    NGSIM Lankershim && 1.726 $\pm$ 0.552  &  1.381 $\pm$ 0.742 &  0.114 $\pm$ 0.051 &  0.159 $\pm$ 0.000 & &   0.324 $\pm$ 0.273 \\
    Stanford Death Circle && 0.453 $\pm$ 0.409  & 0.387 $\pm$ 0.352 &  0.119 $\pm$ 0.044 &  0.159 $\pm$ 0.000  & & 2.344 $\pm$ 5.615\\
    \bottomrule
    \end{tabular}
  \label{table:comp}
\end{table*}

\begin{table}[]
  \centering
  \ra{1}
  \caption{Percentage average improvement of likelihood after combining observations with the prior.}
    \begin{tabular}{@{}lcc@{}}
    \toprule
    \multicolumn{1}{c}{\bf Quantity} & \multicolumn{1}{c}{\bf Roundabout} & \multicolumn{1}{c}{\bf Intersection}\\
    \midrule
    Prior likelihood & 267.072 & 1288.951 \\
    Observation likelihood  & 187.271 & 1051.696 \\
    Posterior likelihood  & 280.905 & 1619.642 \\
    Percentage improvement & 54.119\% & 48.993\%\\
    \bottomrule
    \end{tabular}
  \label{table:comp_impr}
\end{table}

On average, it takes 82 milliseconds for the EM algorithm to learn the parameters of each mixture of von Mises distributions. Using 10\% of the dataset as the test set, to assess how well the distribution is fitted, the average probability density was computed (Table~\ref{table:comp}). Since the test dataset was prepared randomly, it is expected that a higher number of data points are under the higher density area. Hence, the average probability density should be higher than a benchmark uniform distribution. The lower the concentrations of the fitted distribution are, the higher the average density values would be.  However, since this is a density estimation problem, designing a metric for evaluating the dispersion is challenging. We compared our results with DGM, Gaussian process directional estimation (GP) \cite{o2011learning}, and an uninformative von Mises distribution ($\kappa \simeq 0$). We considered the sparse subset of data approximation to GPs \cite{herbrich2003fast} as the datasets we consider are extremely large to fit a full GP in a reasonable time. The last column in Table~\ref{table:comp} also reports average density for the gamma speed distributions.

To assess how much accuracy is gained by combining prior and current belief distributions, with reference to the example given in Figure~\ref{fig:prod}, the RMSE value between the true ($-90^\circ$) and 100 angles sampled from each distribution were computed. The RMSE values of angles dropped from approximately $60^\circ$ to $40^\circ$ after combining prior and current estimates. As reported in Table~\ref{table:comp_impr}, we further evaluated the effectiveness of combining prior information with observations on the Carla roundabout and intersection datasets using the improvement in likelihood as a metric. Firstly, we randomly picked several locations in the environment. We then computed the prior likelihood $\mathcal{L}_0$ on the test set. As a proxy, we considered the direction of observation as the most probable direction indicated by the directional distribution in the cell right before the prior cell. Its likelihood $\mathcal{L}_t$ is also calculated. We then computed the prior-observation combined distribution using eq.~(\ref{eq:post}) and the posterior likelihood $\mathcal{L}_*$ on the test dataset. The percentage improvement in likelihood is $(\mathcal{L}_* - \mathcal{L}_t)/\mathcal{L}_t \times 100\%$. 

We also simulated future trajectories of vehicles. With the fitted distributions, in Figure~\ref{fig:carla2_a}, three possible trajectories starting at three different locations around the roundabout are shown in black arrows. Although only one possible trajectory for each starting location is shown, it is possible to generate all possible trajectories a vehicle could follow using the sampling procedure introduced in Algorithm~\ref{algo:traj}. 

\section{Discussion}
\label{sec:diss}

In Section~\ref{sec:dir_prim}, we divided the environment into cells. This process sometimes splits the important areas of the road among several cells and disregards certain aspects of the underlying road network structure, limiting the accuracy of predictions and increasing the memory requirements to store the parameters. Therefore, it is important to develop an automatic tessellation technique that takes the road geometry into consideration. For that purpose, in future work, as in the context of decentralized collision avoidance \cite{WangEtAlICRA18BICLaneChange} and multi-robot coverage \cite{pierson2017adapting}, it is possible to use Voronoi cells. In $2$D space, a Voronoi diagram is a partition of the space based on the positions of a group of $N$ generators $\{\mathbf{x}_{1}, \mathbf{x}_{2}, \ldots, \mathbf{x}_{N}\}$. The Voronoi cell associated with the $i$th generator consists of all the points that are closer to $\mathbf{x}_{i}$ than any other generators. If Voronoi cell $i$ and $j$ are adjacent, generator $i$ and $j$ are called Voronoi neighbors. The construction of a Voronoi cell only depends on the position of ego agent and its Voronoi neighbors. Thus, the computation of Voronoi partition is fully decentralized and the computation time is linear with respect to the number of neighbors. For the directional map, we can jointly learn the cell boundaries and directional parameters by solving a variant of $\min \sum_{c\in\mathcal{C}} \sum_{\mathbf{x}\in\mathcal{A}} -\log p(\mathbf{x}_c\mid\theta_c)$ for $\mathcal{C}$ cells and $\mathcal{A}$ all possible assignments of data points to cells.

Rather than considering cells are independent, it is possible to learn the interactions between cells. For instance, with reference to Figure~\ref{fig:motivation}, if a vehicle is observed moving straight in the main road, then it can be inferred that the probability a vehicle comes from the two side roads is relatively low. We can also consider the temporal dependencies among cells. For example, similar to the spatial priors discussed in Section~\ref{sec:curr_know}, we can compute temporal priors. Such a formulation paves our way to directional-speed Bayesian filtering \cite{thrun2000probabilistic}.

\section{Conclusion} 
\label{sec:conclusion}
We introduced directional primitives as a framework to represent uncertainty in directions and speeds of a road network. We showed that this prior information can be combined with current information about the vehicles observed in the environment to infer the possible directions a vehicle could turn. Future work will focus on optimizing road tessellation, incorporating spatiotemporal state dependencies, and employing computer vision models for estimating the distribution of observed vehicles. These efforts will lead to developing safe-decision-making algorithms for autonomous vehicles operating in urban environments \cite{sun2019behavior,shu2020autonomous}.

\section*{Acknowledgments}
The authors thank Dr. Alex Koufos for discussions related to the Carla simulator. Toyota Research Institute (TRI) provided funds to assist the authors with their research, but this article solely reflects the opinions and conclusions of its authors and not TRI or any other Toyota entity. 

\renewcommand*{\bibfont}{\small}
\printbibliography

@String { cvpr        = {IEEE Computer Society Conference on Computer Vision and Pattern Recognition (CVPR)} }

@String { icra        = {IEEE International Conference on Robotics and Automation (ICRA)} }

@String { ijrr        = {International Journal of Robotics Research (IJRR)} }

@String { iros        = {IEEE/RSJ International Conference on Intelligent Robots and Systems (IROS)} }

@String { itsc        = {IEEE International Conference on Intelligent Transportation Systems (ITSC)} }

@String { iv          = {IEEE Intelligent Vehicles Symposium (IV)} }

@String { mit         = {Massachusetts Institute of Technology} }

@String { nips        = {Advances in Neural Information Processing Systems (NIPS)} }

@String { rss         = {Robotics: Science and Systems (RSS)} }

@String { uai         = {Conference on Uncertainty in Artificial Intelligence (UAI)} }

@String { ral         = {Robotics and Automation Letters (RA-L)} }

@String { corl         = {Conference on Robot Learning (CoRL)} }

@article{fox2003bayesian,
  title={Bayesian filtering for location estimation},
  author={Fox, Dieter and Hightower, Jeffrey and Liao, Lin and Schulz, Dirk and Borriello, Gaetano},
  journal={IEEE Pervasive Computing},
  volume={2},
  number={3},
  pages={24--33},
  year={2003},
}

@article{tahboub2006intelligent,
  title={Intelligent human-machine interaction based on dynamic {B}ayesian networks probabilistic intention recognition},
  author={Tahboub, Karim A},
  journal={Journal of Intelligent and Robotic Systems},
  volume={45},
  number={1},
  pages={31--52},
  year={2006},
}

@article{polit1979characteristics,
  title={Characteristics of motor programs underlying arm movements in monkeys},
  author={Polit, A and Bizzi, Emilio},
  journal={Journal of Neurophysiology},
  volume={42},
  number={1},
  pages={183--194},
  year={1979},
}

@article{flash2005motor,
  title={Motor primitives in vertebrates and invertebrates},
  author={Flash, Tamar and Hochner, Binyamin},
  journal={Current Opinion in Neurobiology},
  volume={15},
  number={6},
  pages={660--666},
  year={2005},
  publisher={Elsevier}
}

@inproceedings{paraschos2013probabilistic,
  title={Probabilistic movement primitives},
  author={Paraschos, Alexandros and Daniel, Christian and Peters, Jan R and Neumann, Gerhard},
  booktitle=nips,
  pages={2616--2624},
  year={2013}
}

@article{perk2006motion,
  title={Motion primitives for robotic flight control},
  author={Perk, Baris E and Slotine, Jean-Jacques E},
  journal={arXiv preprint cs/0609140},
  year={2006}
}

@inproceedings{lacaze1998path,
  title={Path planning for autonomous vehicles driving over rough terrain},
  author={Lacaze, Alberto and Moscovitz, Yigal and DeClaris, Nicholas and Murphy, Karl},
  booktitle={IEEE International Symposium on Intelligent Control (ISIC)},
  pages={50--55},
  year={1998},
}

@article{pivtoraiko2009differentially,
  title={Differentially constrained mobile robot motion planning in state lattices},
  author={Pivtoraiko, Mihail and Knepper, Ross A and Kelly, Alonzo},
  journal={Journal of Field Robotics},
  volume={26},
  number={3},
  pages={308--333},
  year={2009},
  publisher={Wiley}
}

@inproceedings{campbell2019,
  title={Probabilistic Multimodal Modeling for Human-Robot Interaction Tasks},
  author={Campbell, Joseph and Stepputtis, Simon and Amor, Heni Ben},
  booktitle=rss,
  year={2019}
}

@InProceedings{dasgupta_cvpr16,
  author       = {Alexandre Alahi and  Kratarth Goel and Vignesh Ramanathan and Alexandre Robicquet and Li Fei-Fei and Silvio Savarese},
  title        = {Social {LSTM}: Human Trajectory Prediction in Crowded Spaces},
  booktitle    = cvpr,
  year         = {2016}
}

@inproceedings{park_2018,
    author = {Seong Hyeon Park and ByeongDo Kim and Chang Mook Kang and Chung Choo Chung and Jun Won Choi},
    title = {Sequence-to-Sequence Prediction of Vehicle Trajectory via {LSTM} Encoder-Decoder Architecture},
    pages = {1672--1678},
    booktitle = iv,
    year={2018}
}

@article{wilko_2017,
    author = {Wilko Schwarting and Javier Alonso-Mora and Liam Paull and Sertac Karaman and Daniela Rus},
    title = {Safe Nonlinear Trajectory Generation for Parallel Autonomy with a Dynamic Vehicle Model},
    journal = itsc,
    volume={19},
    number={9},
    pages={2994--3008},
    year={2017},
}

@INPROCEEDINGS{rans_2016,
  author = {Senanayake, Ransalu and Ott, Lionel and O'Callaghan, Simon and Ramos, Fabio},
  title = {Spatio-Temporal Hilbert Maps for Continuous Occupancy Representation in Dynamic Environments},
  booktitle = nips,
  year = {2016}
}

@INPROCEEDINGS{auto_2018,
  author = {Senanayake, Ransalu and Tompkins, Anthony and Ramos, Fabio},
  title = {Automorphing Kernels for Nonstationarity in Mapping Unstructured Environments},
  booktitle = corl,
  year = {2018}
}

@INPROCEEDINGS{bhm_2017,
  author = {Senanayake, Ransalu and Ramos, Fabio},
  title = {Bayesian Hilbert Maps for Dynamic Continuous Occupancy Mapping},
  booktitle = corl,
  year = {2017}
}

@inproceedings{senanayake2018directional,
  title={Directional grid maps: modeling multimodal angular uncertainty in dynamic environments},
  author={Senanayake, Ransalu and Ramos, Fabio},
  booktitle=iros,
  pages={3241--3248},
  year={2018},
}

@article{zhi2019klstm,
  title={Spatiotemporal Learning of Directional Uncertainty in Urban Environments with Kernel Recurrent Mixture Density Networks},
  author={Zhi, William and Senanayake, Ransalu and Ott, Lionel and Ramos, Fabio},
  journal=ral,
  volume={4},
  number={4},
  pages={4306 - 4313},
  year={2019},
}

@inproceedings{kurz2014recursive,
  title={Recursive Estimation of Orientation Based on the {B}ingham Distribution},
  author={Kurz, Gerhard and Gilitschenski, Igor and Julier, Simon and Hanebeck, Uwe D},
  booktitle={International Conference on Information Fusion (ICIF)},
  year=2013
}

@article{gilitschenski2015unscented,
  title={Unscented orientation estimation based on the {B}ingham distribution},
  author={Gilitschenski, Igor and Kurz, Gerhard and Julier, Simon J and Hanebeck, Uwe D},
  journal={IEEE Transactions on Automatic Control},
  volume={61},
  number={1},
  pages={172--177},
  year={2015},
}

@article{MardiaEdw1982,
    author = {Mardia, K. V.  and  Edwards, R.},
    title = {Weighted distributions and rotating caps},
    journal = {Biometrika},
    volume = {69},
    number = {180},
    pages = {323-330},
    year  = {1982},
}

@book{DirectStat,
  title={Directional Statistics},
  author={Mardia, Kanti V.  and  Jupp, Peter E.},
  year={2006},
  publisher={Wiley}
}

@inproceedings{IvanovicSchmerlingEtAl2018,
  author = {Ivanovic, B. and Schmerling, E. and Leung, K. and Pavone, M.},
  title = {Generative Modeling of Multimodal Multi-Human Behavior},
  booktitle = iros,
  year = {2018},
}

@inproceedings{ester1996density,
  title={A density-based algorithm for discovering clusters in large spatial databases with noise.},
  author={Ester, Martin and Kriegel, Hans-Peter and Sander, J{\"o}rg and Xu, Xiaowei and others},
  booktitle={ACM SIGKDD Conference on Knowledge Discovery and Data Mining (KDD)},
  volume={96},
  number={34},
  pages={226--231},
  year={1996},
}

@inproceedings{vintr2019time,
  title={Time-varying pedestrian flow models for service robots},
  author={Vintr, Tom{\'a}{\v{s}} and Molina, Sergi and Senanayake, Ransalu and Broughton, George and Yan, Zhi and Ulrich, Ji{\v{r}}{\'\i} and Kucner, Tomasz Piotr and Swaminathan, Chittaranjan Srinivas and Majer, Filip and Stachov{\'a}, M{\'a}ria and others},
  booktitle={2019 European Conference on Mobile Robots (ECMR)},
  pages={1--7},
  year={2019},
  organization={IEEE}
}

@inproceedings{vintr2019spatio,
  title={Spatio-temporal representation for long-term anticipation of human presence in service robotics},
  author={Vintr, Tom{\'a}{\v{s}} and Yan, Zhi and Duckett, Tom and Krajn{\'\i}k, Tom{\'a}{\v{s}}},
  booktitle=icra,
  pages={2620--2626},
  year={2019},
  organization={IEEE}
}

@article{anderson1975improved,
  title={Improved maximum likelihood estimators for the gamma distribution},
  author={Anderson, CW and Ray, WD},
  journal={Communications in Statistics-Theory and Methods},
  volume={4},
  number={5},
  pages={437--448},
  year={1975},
  publisher={Taylor \& Francis}
}

@article{ulrich1984computer,
  title={Computer Generation of Distributions on the M-Sphere},
  author={Ulrich, Gary},
  journal={Journal of the Royal Statistical Society: Series C (Applied Statistics)},
  volume={33},
  number={2},
  pages={158--163},
  year={1984},
  publisher={Wiley Online Library}
}

@article{s-vae18,
  title={Hyperspherical Variational Auto-Encoders},
  author={Davidson, Tim R. and
          Falorsi, Luca and
          De Cao, Nicola and
          Kipf, Thomas and
          Tomczak, Jakub M.},
  journal = uai,
  year = {2018}
}

@article{glynn1989importance,
  title={Importance sampling for stochastic simulations},
  author={Glynn, Peter W and Iglehart, Donald L},
  journal={Management science},
  volume={35},
  number={11},
  pages={1367--1392},
  year={1989},
  publisher={INFORMS}
}

@article{casella2004generalized,
  title={Generalized Accept-Reject Sampling Schemes},
  author={Casella, George and Robert, Christian P and Wells, Martin T},
  journal={Lecture Notes-Monograph Series},
  pages={342--347},
  year={2004},
  publisher={JSTOR}
}

@incollection{davis2011remarks,
  title={Remarks on some nonparametric estimates of a density function},
  author={Davis, Richard A and Lii, Keh-Shin and Politis, Dimitris N},
  booktitle={Selected Works of Murray Rosenblatt},
  pages={95--100},
  year={2011},
  publisher={Springer}
}

@book{casella2002statistical,
  title={Statistical inference},
  author={Casella, George and Berger, Roger L},
  volume={2},
  year={2002},
  publisher={Duxbury Pacific Grove, CA}
}

@inproceedings{ihler2004efficient,
  title={Efficient multiscale sampling from products of Gaussian mixtures},
  author={Ihler, Alexander T and Sudderth, Erik B and Freeman, William T and Willsky, Alan S},
  booktitle=nips,
  pages={1--8},
  year={2004}
}

@article{dosovitskiy2017carla,
  title={{CARLA}: An open urban driving simulator},
  author={Dosovitskiy, Alexey and Ros, German and Codevilla, Felipe and Lopez, Antonio and Koltun, Vladlen},
  journal={arXiv preprint arXiv:1711.03938},
  year={2017}
}

@inproceedings{robicquet2016learning,
  title={Learning social etiquette: Human trajectory understanding in crowded scenes},
  author={Robicquet, Alexandre and Sadeghian, Amir and Alahi, Alexandre and Savarese, Silvio},
  booktitle={European Conference on Computer Vision (ECCV)},
  pages={549--565},
  year={2016},
  organization={Springer}
}

@article{alexiadis2004next,
  title={The next generation simulation program},
  author={Alexiadis, Vassili and Colyar, James and Halkias, John and Hranac, Rob and McHale, Gene},
  journal={Journal of the Institute of Transportation Engineers (ITE)},
  volume={74},
  number={8},
  pages={22},
  year={2004},
  publisher={Institute of Transportation Engineers}
}

@inproceedings{o2011learning,
  title={Learning navigational maps by observing human motion patterns},
  author={O'Callaghan, Simon T and Singh, Surya PN and Alempijevic, Alen and Ramos, Fabio T},
  booktitle=icra,
  pages={4333--4340},
  year={2011},
}

@inproceedings{shan2019extended,
  title={Extended Vehicle Tracking with Probabilistic Spatial Relation Projection and Consideration of Shape Feature Uncertainties},
  author={Shan, Mao and De Alvis, Charika and Worrall, Stewart and Nebot, Eduardo},
  booktitle={IEEE Intelligent Vehicles Symposium (IV)},
  pages={1477--1483},
  year={2019},
}

@inproceedings{WangEtAlICRA18BICLaneChange,
  author = {M. Wang and Z. Wang and S. Paudel and M. Schwager}, 
  title = {Safe Distributed Lane Change Maneuvers for Multiple Autonomous Vehicles Using Buffered Input Cells},
  booktitle = icra,
  month = {5},
  year = {2018},
  pages = {4678--4684}
}

@article{pierson2017adapting,
  title={Adapting to sensing and actuation variations in multi-robot coverage},
  author={Pierson, Alyssa and Figueiredo, Lucas C and Pimenta, Luciano CA and Schwager, Mac},
  journal=ijrr,
  volume={36},
  number={3},
  pages={337--354},
  year={2017},
  publisher={SAGE Publications Sage UK: London, England}
}

@inproceedings{herbrich2003fast,
  title={Fast sparse Gaussian process methods: The informative vector machine},
  author={Herbrich, Ralf and Lawrence, Neil D and Seeger, Matthias},
  booktitle=nips,
  pages={625--632},
  year={2003}
}

@book{thrun2000probabilistic,
  title={Probabilistic Robotics},
  author={Thrun, Sebastian and Burgard, Wolfram and Fox, Dieter},
  year={2000},
  publisher={MIT Press}
}

@inproceedings{sun2019behavior,
  title={Behavior planning of autonomous cars with social perception},
  author={Sun, Liting and Zhan, Wei and Chan, Ching-Yao and Tomizuka, Masayoshi},
  booktitle={IEEE Intelligent Vehicles Symposium (IV)},
  pages={207--213},
  year={2019},
}

@article{shu2020autonomous,
  title={Autonomous Driving at Intersections: A Critical-Turning-Point Approach for Left Turns},
  author={Shu, Keqi and Yu, Huilong and Chen, Xingxin and Chen, Long and Wang, Qi and Li, Li and Cao, Dongpu},
  journal={arXiv preprint arXiv:2003.02409},
  year={2020}
}

\end{document}